\documentclass[10pt,twocolumn,letterpaper]{article}

\usepackage{cvpr}              
\definecolor{cvprblue}{rgb}{0.21,0.49,0.74}
\usepackage[pagebackref,breaklinks,colorlinks,allcolors=cvprblue]{hyperref}
\usepackage{wrapfig,algorithm,amssymb,algpseudocode}
\usepackage{multirow}
\usepackage{multicol}
\usepackage[table]{xcolor}
\definecolor{myblue}{rgb}{0.88, 0.95, 0.99}
\definecolor{bluecomm}{rgb}{0.25, 0.4, 0.55}

\def\paperID{****} 
\def\confName{CVPR}
\def\confYear{2026}

\title{AwareVLN: Reasoning with Self-awareness for Vision-Language Navigation}

\author{
Wenxuan Guo \quad 
Xiuwei Xu\textsuperscript{$*$} \quad
Yichen Liu \quad
Xiangyu Li \quad
Hang Yin \quad
Huangxing Chen \\
Wenzhao Zheng \quad
Jianjiang Feng\textsuperscript{$\dagger$} \quad
Jie Zhou \quad
Jiwen Lu\textsuperscript{$\dagger$} \\
Tsinghua University, China
}

\begin{document}

\twocolumn[{%
\renewcommand\twocolumn[1][]{#1}%
\maketitle
\begin{center}
    \centering
    \vspace{-.4cm}
    \captionsetup{type=figure}
    \includegraphics[width=1.0\textwidth]{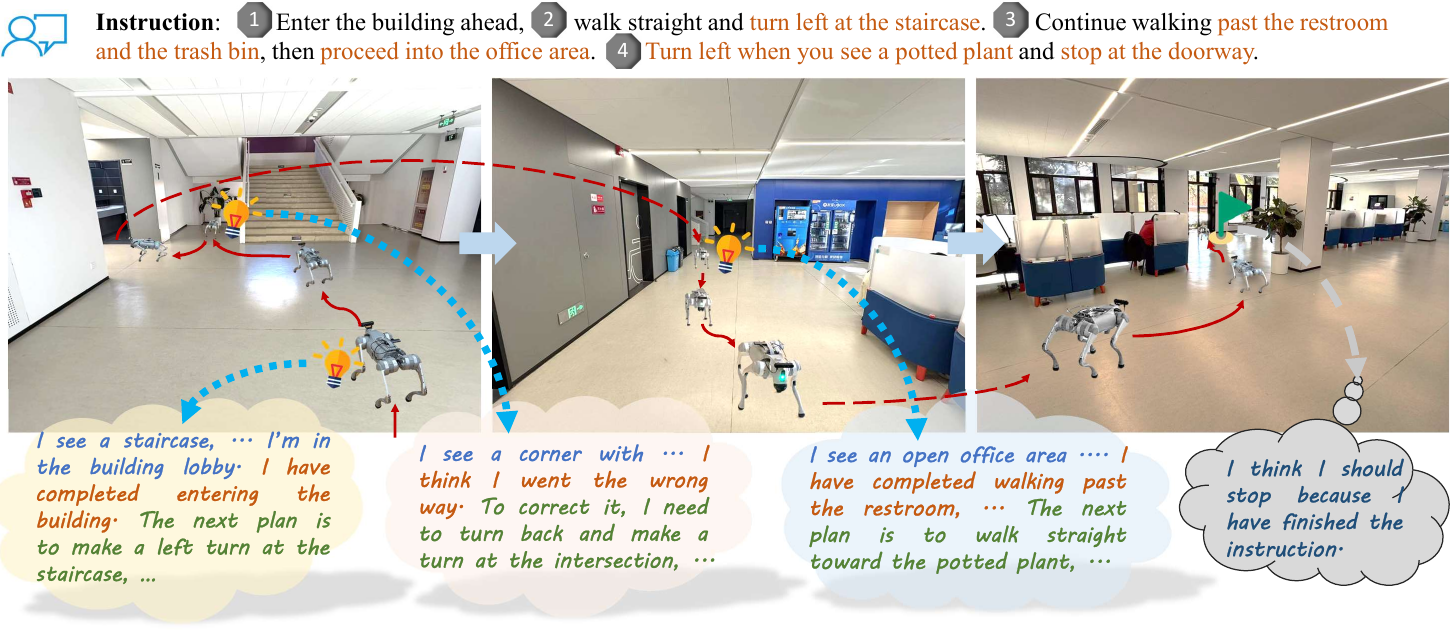}
    \vspace{-.6cm}
    \caption{AwareVLN equips a VLN agent with self-aware, structured reasoning that is selectively triggered at key navigation points. Instead of relying solely on end-to-end action prediction, AwareVLN enables the agent to explicitly analyze its spatial state, task progress, and alignment with the instruction when such reasoning is truly needed, achieving more robust and explainable instruction following.}
    \label{fig:teaser}
\end{center}
}]

{
        \renewcommand{\thefootnote}{* \,}
	\footnotetext[5]{Project leader. \, $^{\dagger}$ Corresponding author.}
}

\begin{abstract}
    
Vision-and-Language Navigation (VLN) requires an agent to ground language instructions to its own movement within a visual environment. While state-of-the-art methods leverage the reasoning capabilities of Vision-Language Models (VLMs) for end-to-end action prediction, they often lack an explicit and explainable understanding of the relationships between the agent, the instruction, and the scene. Conversely, explicitly building a scene map for heuristic planning is intuitively appealing but relies on additional 3D sensors and hinders large-scale vision-language pre-training. To bridge this gap, we propose AwareVLN, a novel framework that equips the navigation model with a self-aware reasoning mechanism, enabling it to understand the agent's state and task progress in a fully end-to-end and data-driven manner. Our approach features two key innovations: (1) a structural reasoning module that fosters spatial and task-oriented self-awareness, and (2) an automatic data engine with progress division for effective training. Extensive experiments on various datasets in Habitat simulator show our AwareVLN significantly outperforms previous state-of-the-art vision-language navigation methods. 
Project page: \href{https://gwxuan.github.io/AwareVLN/}{https://gwxuan.github.io/AwareVLN/}.
 
\end{abstract}

\section{Introduction}
    Vision-Language Navigation (VLN)~\cite{anderson2018vision} requires an agent to navigate in an unknown environment following natural language instructions, grounding linguistic concepts to visual scenes and physical movements. Conventional approaches~\cite{wang2023dreamwalker,an2024etpnav} typically rely on constructing explicit topological graphs of the environment, upon which they perform task planning and vision-language grounding. While effective, this paradigm often depends on precise 3D sensing and SLAM systems. Recently, end-to-end Vision-Language-Model-based methods~\cite{zhang2024navid,zhang2024uninavid,cheng2024navila} have emerged as a promising alternative, eliminating the need for additional sensors by directly mapping instructions and observations to robot actions through large-scale vision-language pre-training and vision-language-action co-training.

However, current VLM-based VLN methods primarily focus on taming VLMs for direct action prediction, overlooking the potential to harness their inherent reasoning capabilities. Consequently, the resulting end-to-end navigation process remains largely unexplainable and lacks robustness, struggling with precise subtask planning and error correction due to a fundamental lack of self-awareness. 
Nav-R1~\cite{liu2025nav} attempts to explicitly reason through a dual-system mechanism at fixed intervals during navigation.
However, its reasoning supervision data are derived from a generic VLM queried on past observations, lacking rich self-aware knowledge. As a result, its reasoning often fails to provide deep insights into the navigation process and the occurred trajectory errors, and its reasoning merely serving as a text output rather than guiding subsequent action generation.
Therefore, how to accurately reason about current agent's state and task progress according to observation history remains an underexplored question.

In this paper, we propose AwareVLN, a novel self-aware reasoning framework for VLN that moves beyond pure action prediction or simplistic thinking generation. Unlike existing VLM-based methods that either predict only actions or jointly output actions with corresponding explanations, AwareVLN introduces a sparse reasoning mechanism that performs structured, in-depth analysis of the agent's relationship with the instruction and environment only at key navigation nodes. This design ensures both computational efficiency and genuine self-awareness by strategically triggering reasoning when most beneficial.
Specifically, our framework incorporates two key technical innovations: (1) A structural reasoning architecture that enables the model to autonomously decide when to engage in reasoning. When activated, it thoroughly analyzes the current agent-instruction-environment relationship by synthesizing past visual observations and previous reasoning results; (2) An automatic data engine with progress-aware division strategy that systematically identifies and breaks down key navigation milestones, generating targeted high-quality training data to effectively teach the model about task progress analysis and high-level planning.
Extensive experiments on various datasets in Habitat simulator show our AwareVLN significantly outperforms previous state-of-the-art vision-language navigation methods.

\section{Related Work}
    \textbf{Vision-and-language Navigation.}
Early work in vision-and-language navigation (VLN)~\cite{liu2023bird,long2024discuss} primarily employed discrete simulators, where environments are represented as graphs of navigable waypoints. Agents navigate by sequentially selecting from adjacent nodes. This abstraction simplifies the problem by focusing on high-level path planning, but the significant sim-to-real gap hinders deployment in physical settings.
To bridge this gap, Vision-and-Language Navigation in Continuous Environments (VLN-CE)~\cite{krantz_vlnce_2020} was introduced, enabling agents to move freely via low-level action predictions. Subsequent methods~\cite{krantz2022sim, wang2023dreamwalker, wang2023gridmm, Wang_lookahead, an2024etpnav} designed for VLN-CE can be directly transferred to real-world robots. A common practice is to use waypoint predictor~\cite{Hong_2022_CVPR}, which predicts discrete waypoints online in a continuous space and bridge the continuous and discrete environments.
Recently, end-to-end Vision-Language Models (VLMs)-based methods~\cite{zhang2024navid,zhang2024uninavid,cheng2024navila,yu2025correctnav} have emerged. These methods directly map language instruction and visual observations to actions, enhancing robustness by eliminating dependencies on 3D sensors or SLAM system, and demonstrating improved generalization thanks to large-scale pretraining. However, they typically predict low-level actions without reasoning about the agent's state or task progress, limiting their ability to recover from errors or perform nuanced planning. Our AwareVLN addresses this by fully leveraging the reasoning capabilities of VLMs for self-aware navigation.

\noindent\textbf{Reasoning in Embodied Model.}
In embodied AI, leveraging LLM/VLM reasoning beyond low-level action prediction is crucial for explainable and robust systems. In manipulation, methods are broadly categorized as integrated or hierarchical~\cite{gao2025vla}. Integrated methods use a single model for planning and policy: EmbodiedCoT~\cite{zawalski2024robotic} and CotVLA~\cite{zhao2025cot} generate language or image chain-of-thought~\cite{wei2022chain} before acting; OneTwoVLA~\cite{lin2025onetwovla} triggers reasoning only at critical steps; MDT~\cite{reuss2024multimodal} and PIDM~\cite{tian2024predictive} employ auxiliary objectives like goal foresight, while RoboBrain~\cite{ji2025robobrain} and ChatVLA~\cite{zhou2025chatvla} use language-based reasoning losses. Hierarchical methods separate planning and policy: RT-H~\cite{belkhale2024rt} uses two VLMs for language and action generation, a structure followed by others with varied architectures~\cite{shi2025hi,wen2025dexvla} or multimodal plans like affordance~\cite{gao2024flip,nasiriany2025rt,wen2023any} and future videos~\cite{du2023learning,yang2023learning}.
However, such reasoning remains scarce in navigation. 
A recent work Nav-R1~\cite{liu2025nav} attempts to explicitly reason through a dual-system mechanism at fixed intervals, but its reasoning supervision data are derived from a generic VLM queried on past observations, lacking rich self-aware knowledge.
In contrast, our AwareVLN leverages a progress-aware data generation engine to implement self-aware reasoning at key navigation nodes, thereby achieving both high efficiency and strong performance.

\section{Method}
    \subsection{Problem Formulation}
\begin{figure*}
	\centering
	\includegraphics[width=0.95\linewidth]{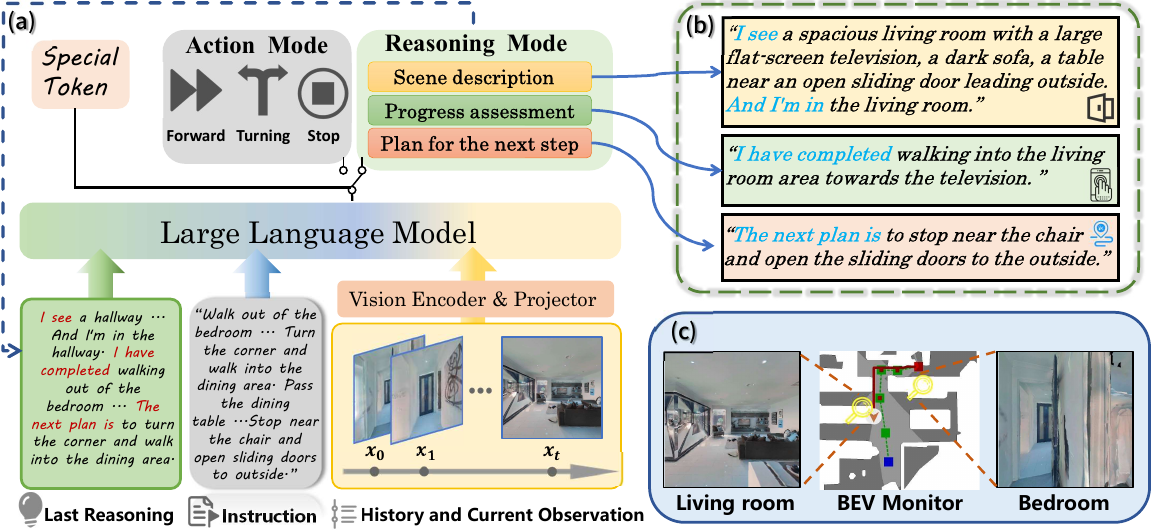}
    \vspace{-.1cm}
    \caption{Framework of AwareVLN. (a) AwareVLN equips a unified vision-language model with both action prediction and self-reflective reasoning, allowing the agent to leverage past reasoning to guide future decisions. (b) The reasoning process is multi-dimensional and causal, sequentially describing the current scene, assessing progress, and planning the next step.(c) As illustrated by the BEV monitor, reasoning is sparsely and structurally triggered at key nodes, such as subtask boundaries.}
	\label{fig:method1}
	\vspace{-.2cm}
\end{figure*}

In this work, we consider vision-and-language navigation in unknown and continuous 3D environments (VLN-CE). 
The agent is provided with a natural-language instruction 
$\mathcal{I}=\{w_1, \dots, w_l\}$, which describes how to navigate from the starting point to the destination, as illustrated in Figure~\ref{fig:method1}. 
During execution, at each time step $t$, the agent receives an egocentric RGB observation stream 
$\mathcal{O}_t=\{\mathbf{x}_0, \dots, \mathbf{x}_t\}$ captured by a monocular camera, 
without access to any depth or pose information.
The goal is to follow the instruction $\mathcal{I}$ by predicting a sequence of executable low-level actions from a discrete action set 
$\mathcal{A}=\{\texttt{FORWARD}, \texttt{TURN-LEFT}, \texttt{TURN-RIGHT}, \texttt{STOP}\}$, 
where each action represents a motion primitive (e.g., moving $25$\,cm forward or rotating $15^\circ$), as commonly adopted in embodied navigation~\cite{savva2019habitat,chaplot2020object}. 
At each step, the model takes $(\mathcal{I}, \mathcal{O}_t)$ as input and produces a textual prediction describing the next movement (e.g., ``move forward 75\,cm''), 
which is then parsed into one or more low-level actions $\mathbf{a}_{t+1:t+k}$ for execution, allowing the agent to navigate effectively in unseen environments.

\subsection{Framework of AwareVLN}
Existing VLN models often rely on direct mappings from visual observations and language instructions to actions, yet they lack \emph{self-aware reasoning}—the ability to assess navigation progress, detect mistakes, and adaptively correct them.
When faced with complex or ambiguous instructions, the agent is required to reason about its own progress and make adaptive decisions. To this end, we introduce \textbf{AwareVLN}, which enables the model to perform self-aware reasoning for understanding navigation context, judging current progress, identifying errors, and generating next plans.
AwareVLN introduces a unified vision-language model (VLM) that jointly handles both action prediction and self-reflective reasoning during navigation. Compared to using two separate models, this unified architecture enables the model to internalize knowledge from both reasoning and acting dimensions, allowing them to interact and mutually enhance each other.

\textbf{Unified Reason-act Framework.}
AwareVLN adopts a unified vision-language framework that jointly performs reasoning and action prediction. 
Given the navigation instruction $\mathcal{I}$ and the sequence of visual observations $\mathcal{O}_t=\{\mathbf{x}_0,\dots,\mathbf{x}_t\}$, 
a tokenizer $f_{\mathrm{tok}}(\cdot)$ is used to convert textual inputs (i.e., the instruction $\mathcal{I}$ and the most recent reasoning output $\mathcal{R}$) into token sequences, 
while a vision encoder $f_{\mathrm{vis}}(\cdot)$ extracts visual embeddings from raw RGB observations. For efficiency, we uniformly sample $8$ frames of observations as the visual input.
The number of steps between the current frame and the previous reasoning step is additionally encoded as a relative positional cue to facilitate temporal grounding, 
and fused with the reasoning text to provide explicit temporal context:
\begin{equation}
\mathcal{R}' = \mathcal{R} \oplus (t - t_{\mathrm{prev}}),
\end{equation}
where $t_{\mathrm{prev}}$ denotes the time step of the last reasoning output. Based on this combined context, the unified policy $\pi_{\theta}$ produces a logit $d$ to decide special token $\mathcal{D}$ and a textual output $y_t$:
\begin{equation}
d, y_t = \pi_{\theta}\big(
f_{\mathrm{tok}}(\mathcal{I}),\,
f_{\mathrm{tok}}(\mathcal{R}'),\,
f_{\mathrm{vis}}(\mathcal{O}_t)
\big).
\end{equation}
\begin{equation}
\mathcal{D} =
\begin{cases}
\texttt{[REASON]}, & \text{if } d_{\texttt{[REASON]}} > d_{\texttt{[ACT]}},\\[4pt]
\texttt{[ACT]}, & \text{otherwise.}
\end{cases}
\end{equation}
As illustrated in Algorithm~\ref{algo:awarevln_pipeline}, when \texttt{[REASON]} is predicted, the model enters the reasoning mode and produces a textual description that summarizes the agent’s understanding and progress.
Conversely, when \texttt{[ACT]} is predicted, the model switches to the acting mode and generates an action command specifying the next executable steps.

This unified, language-driven formulation allows AwareVLN to integrate perception, reasoning, and control seamlessly. By recurrently conditioning on previous reasoning and relative step information, the model maintains temporal awareness and achieves adaptive decision-making across complex, long-horizon navigation tasks.
\begin{figure}[t]
  \centering
  \begin{minipage}{0.95\linewidth}
  \begin{algorithm}[H]
  \small
  \begin{algorithmic}[1]
  \Require Policy $\pi_{\theta}$, instruction $\mathcal{I}$, initial observation $\mathbf{x}_0$
  \State $t \gets 0$;\quad $\mathcal{R} \gets$ ``None'';\quad $t_{\mathrm{prev}} \gets 0$
  \While{task not finished}
    \State $\mathcal{R}' = \mathcal{R} \oplus (t - t_{\mathrm{prev}})$ \Comment{\textcolor{bluecomm}{Fuse relative step distance with reasoning text}}
    \State $d,\, y_t = \pi_{\theta}\big(
        f_{\mathrm{tok}}(\mathcal{I}),\,
        f_{\mathrm{tok}}(\mathcal{R}),\,
        f_{\mathrm{vis}}(\mathcal{O}_t)
      \big)$ \Comment{\textcolor{bluecomm}{Output the first logit and text content}}
    \State $\mathcal{D} \gets \text{token from } \max(d_{\texttt{[REASON]}},d_{\texttt{[ACT]}})$ \Comment{\textcolor{bluecomm}{Determine special mode token}}
    \If{$\mathcal{D} = \texttt{[REASON]}$}
        \State $\mathcal{R} \gets y_t$ \Comment{\textcolor{bluecomm}{Update last reasoning}}
        \State $t_{\mathrm{prev}} \gets t$ \Comment{\textcolor{bluecomm}{Record the time of last reasoning}}
    \ElsIf{$\mathcal{D} = \texttt{[ACT]}$}
        \State $\mathbf{a}_{t+1:t+k} \gets \textsc{Parse}(y_t)$ \Comment{\textcolor{bluecomm}{Parse action output into low-level actions}}
        \State \textsc{Execute}$(\mathbf{a}_{t+1:t+k})$
    \EndIf
    \State $t \gets t + 1$
    \State $\mathcal{O}_t \gets \mathcal{O}_{t-1} \cup \{\mathbf{x}_t\}$ \Comment{\textcolor{bluecomm}{Append new frame to the observation sequence}}
  \EndWhile
  \end{algorithmic}
  \caption{Navigation pipeline of AwareVLN with unified, language-driven reasoning and acting.}
  \label{algo:awarevln_pipeline}
  \end{algorithm}
  \end{minipage}
  \vspace{-.4cm}
\end{figure}

\textbf{Structural Reasoning for Self-awareness.}
Accurately following navigation instructions in complex environments requires the agent to maintain a clear understanding of its own progress. Motivated by this observation, we design a structured reasoning mechanism that enables the model to explicitly reason about its navigation status at critical points rather than continuously at every step. This design not only allows the agent to analyze its self-navigation progress and identify potential deviations, but also ensures reasoning efficiency by avoiding unnecessary inference interruptions.

Our AwareVLN will trigger reasoning at key states including: (\textit{i}) \emph{Subtask completion.} When the model detects that a sub-instruction (e.g., ``walk to the doorway'') has been fulfilled, it triggers reasoning to summarize the current progress, confirm the subgoal completion, and plan the next step toward the global goal. This helps the agent maintain coherence across multiple instruction segments., (\textit{ii}) \emph{Path deviation.} When the model identifies inconsistencies between the expected and observed visual cues—such as missing landmarks or incorrect spatial alignment—it enters reasoning mode to analyze potential navigation errors and proposes corrective actions to realign the route., and (\textit{iii}) \emph{Stopping error.} In the final stage of navigation, when the agent detects that its current visual context deviates from the target description, it triggers reasoning to analyze the possible discrepancy and adjust the subsequent plan. This reflective process helps the agent achieve precise goal localization and avoid terminal errors.

We introduce a triplet-based structural reasoning format as shown in Figure~\ref{fig:method1}(b), to endow the agent with explicit self-awareness. AwareVLN is required to perform structured reasoning through the following components:
\begin{itemize}
    \item[(1)] \textbf{Scene description:} a concise depiction of the visual context and surrounding environment at the key node;  
    \item[(2)] \textbf{Progress assessment:} an analysis of the current status, including which parts of the instruction have been completed and whether any deviations have occurred;  
    \item[(3)] \textbf{Plan for the next step:} a high-level intention or strategy for the next navigation phase.  
\end{itemize}

This structured reasoning design provides explicit self-awareness for the model, bridging perception, reasoning, and planning in a unified language space.

\begin{figure*}
	\centering
	\includegraphics[width=0.97\linewidth]{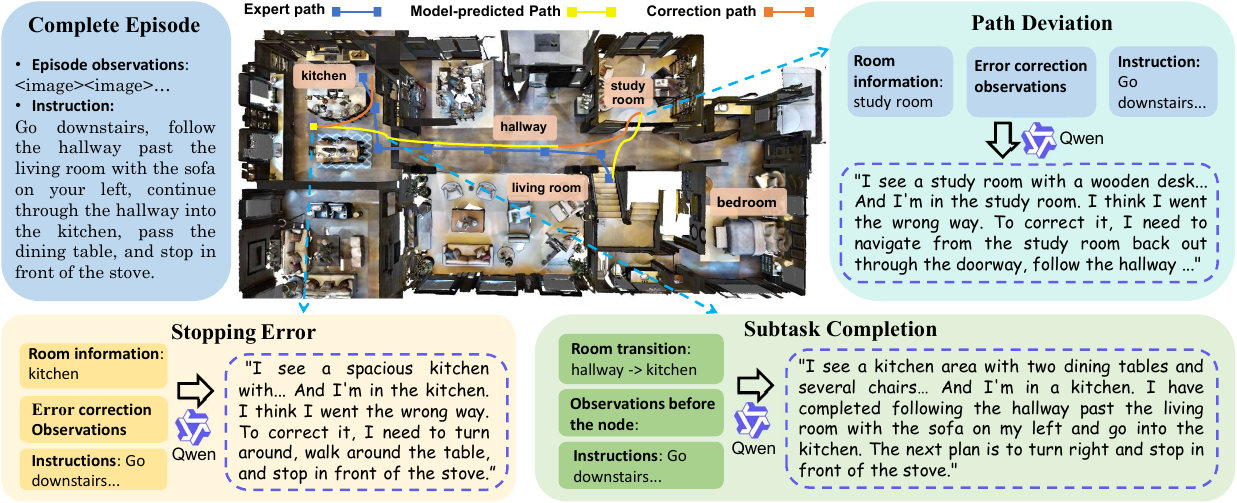}
    \vspace{-.1cm}
    \caption{Automatic Data Engine. Key reasoning nodes are automatically identified using room-level semantics and ground-truth waypoints in the simulator, covering key events including subtask completion, path deviation, and incorrect stopping. For each key node, rich multimodal context is extracted and fed into a general VLM to automatically generate structured, causal reasoning supervision. This pipeline enables scalable, annotation-free construction of high-quality reasoning data.}
	\label{fig:method2}
	\vspace{-.2cm}
\end{figure*}

\subsection{Automatic Data Engine}
To efficiently obtain large-scale and high-quality supervision for reasoning, we design an automatic data engine without any manual annotation. Specifically, we first collect navigation trajectories in the Habitat simulator, where room-level semantic annotations and ground-truth waypoints are utilized to automatically determine key reasoning nodes such as subtask completions, route deviations, and terminal offsets. For each identified key node, we record comprehensive contextual information. Subsequently, this rich navigation context is fed into a general VLM to automatically generate structured reasoning supervision. This process enables scalable acquisition of high-quality reasoning data aligned with real navigation progress, laying a solid foundation for training self-aware navigation models.

\textbf{Navigation Data Collection.}
Constructing navigation data for self-aware supervision involves two components: collecting navigation trajectories and identifying key reasoning nodes. We describe each in detail below.

To collect diverse and informative navigation trajectories, we adopt two complementary data collection strategies. The first is \textit{ground-truth following}, where the agent strictly follows the reference trajectory provided in VLN datasets. This approach ensures highly accurate and instruction-aligned trajectories, serving as ideal samples for modeling correct reasoning behavior. The second is \textit{DAgger-based collection}, where we use an early-stage VLN model—trained solely on expert trajectories—to navigate in the Habitat simulator by executing its predicted actions. When the agent deviates from the ground-truth route, it is corrected back to the next waypoint according to the reference path. This strategy produces more realistic trajectories containing natural prediction errors and correction behaviors, which are particularly valuable for generating reasoning supervision related to error recognition and recovery.

To identify key reasoning nodes, we leverage the scene-level semantic information provided by the simulator and the annotated waypoints from VLN datasets. In this way, we are able to access rich contextual signals and automatically identify key reasoning-triggered states. Specifically, for \textit{subtask completion}, we observe that such transitions often coincide with room changes in the environment (e.g., “leave the bedroom and turn right into the living room”). Therefore, we determine subtask completion based on changes in room category along the trajectory, and also treat the completion of a correction process as a subtask boundary. For \textit{path deviation} and \textit{stopping error}, we compute the spatial deviation between the executed trajectory and the ground-truth waypoints. When the error exceeds a threshold, the step is marked as a deviation node, and subsequent correction observations are recorded. This automatic detection mechanism allows our data engine to efficiently collect large-scale reasoning supervision aligned with realistic navigation progress and diverse error patterns.

\textbf{Reasoning Supervision Generation.}
To generate high-quality reasoning supervision for navigation, we leverage a general VLM, Qwen-VL-Max~\citep{wang2024qwen2}, to transform the collected contextual information into structured reasoning outputs. The VLM receives sufficient multimodal cues to infer accurate reasoning results. As illustrated in Figure \ref{fig:method2}, we design a multi-turn conversation pipeline that progressively guides the VLM to understand the navigation episode and produce causal, interpretable reasoning supervision.
In the first round, we input the complete episode observation sequence together with the language instruction, allowing the VLM to establish a global understanding of the navigation context. In the subsequent rounds, for each identified key node, we feed the following information into the model: the node type, the downsampled visual observations preceding the node, the corresponding room transition information, and the estimated navigation progress (i.e., the ratio of traveled distance to the total path length). For nodes identified as route deviations, we additionally provide the visual observations from the subsequent correction process, enabling the VLM to infer causal relations between errors and recovery behaviors.

\begin{table*}[t]
    \small
    \centering
    \caption{Comparison with state-of-the-art methods on the Val-Unseen split of R2R-CE~\citep{krantz_vlnce_2020} and RxR-CE~\citep{ku2020room}. $^{*}$ indicates methods using the waypoint predictor from~\citet{Hong_2022_CVPR}. AwareVLN outperforms all methods that do not rely on simulator pre-trained waypoint predictors, even when those methods leverage additional inputs such as depth, panoramic views, and odometry.}
    \vspace{-4pt}
    \setlength{\tabcolsep}{6.7pt}
    \scalebox{0.99}{
{\fontsize{7pt}{8pt}\selectfont
\begin{tabular}{lcccclcccclcccc}
\toprule
& \multicolumn{4}{c}{Observation} & & \multicolumn{4}{c}{R2R Val-Unseen} & & \multicolumn{4}{c}{RxR Val-Unseen} \\
\cmidrule(lr){2-5} \cmidrule(lr){7-10} \cmidrule(lr){12-15}
& S.RGB & Pano. & Depth & Odo. & & NE $\downarrow$ & OS $\uparrow$ & SR $\uparrow$ & SPL $\uparrow$ & & NE $\downarrow$ & SR $\uparrow$ & SPL $\uparrow$ & nDTW $\uparrow$ \\
\midrule
HPN+DN$^{*}$~\citep{krantz2021waypoint} &  & \checkmark & \checkmark & \checkmark & & 6.31 & 40.0 & 36.0 & 34.0 & & - & - & - & - \\
CMA$^{*}$~\citep{Hong_2022_CVPR} & & \checkmark & \checkmark & \checkmark & & 6.20 & 52.0 & 41.0 & 36.0 & & 8.76 & 26.5 & 22.1 & 47.0 \\
VLN$\circlearrowright$BERT$^{*}$~\citep{Hong_2022_CVPR} & & \checkmark & \checkmark & \checkmark & & 5.74 & 53.0 & 44.0 & 39.0 & & 8.98 & 27.0 & 22.6 & 46.7 \\
Sim2Sim$^{*}$~\citep{krantz2022sim} & & \checkmark  & \checkmark & \checkmark & & 6.07 & 52.0 & 43.0 & 36.0 & & - & - & - & - \\
GridMM$^{*}$~\citep{wang2023gridmm} & & \checkmark  & \checkmark & \checkmark & & 5.11 & 61.0 & 49.0 & 41.0 & & - & - & - & - \\
Ego$^{2}$-Map$^{*}$~\citep{hong2023learning} & & \checkmark  & \checkmark & \checkmark & & 5.54 & 56.0 & 47.0 & 41.0 & & - & - & - & - \\
DreamWalker$^{*}$~\citep{wang2023dreamwalker} & & \checkmark  & \checkmark & \checkmark & & 5.53 & 59.0 & 49.0 & 44.0 & & - & - & - & - \\
Reborn$^{*}$~\citep{an20221st} & & \checkmark  & \checkmark & \checkmark & & 5.40 & 57.0 & 50.0 & 46.0 & & 5.98 & 48.6 & 42.0 & 63.3 \\
ETPNav$^{*}$~\citep{an2024etpnav} & & \checkmark  & \checkmark & \checkmark & & 4.71 & 65.0 & 57.0 & 49.0 & & 5.64 & 54.7 & 44.8 & 61.9 \\
HNR$^{*}$~\citep{wang2024lookahead} & & \checkmark  & \checkmark & \checkmark & & 4.42 & 67.0 & 61.0 & 51.0 & & 5.50 & 56.3 & 46.7 & 63.5 \\
BEVBert$^{*}$~\citep{an2022bevbert} & & \checkmark  & \checkmark & \checkmark & & 4.57 & 67.0 & 59.0 & 50.0 & & 4.00 & 68.5 & - & 69.6 \\
HAMT+ScaleVLN$^{*}$~\citep{wang2023scaling} & & \checkmark  & \checkmark & \checkmark & & 4.80 & - & 55.0 & 51.0 & & - & - & - & - \\
\midrule
AG-CMTP~\citep{chen2021topological} & & \checkmark  & \checkmark & \checkmark & & 7.90 & 39.0 & 23.0 & 19.0 & & - & - & - & - \\
R2R-CMTP~\citep{chen2021topological} & & \checkmark  & \checkmark & \checkmark & & 7.90 & 38.0 & 26.0 & 22.0 & & - & - & - & - \\
LAW~\citep{raychaudhuri2021language} & \checkmark & & \checkmark & \checkmark & & 6.83 & 44.0 & 35.0 & 31.0 & & 10.90 & 8.0 & 8.0 & 38.0 \\
CM2~\citep{georgakis2022cross} & \checkmark & & \checkmark & \checkmark & & 7.02 & 41.0 & 34.0 & 27.0 & & - & - & - & - \\
WS-MGMap~\citep{chen2022weakly} & \checkmark & & \checkmark & \checkmark & & 6.28 & 47.0 & 38.0 & 34.0 & & - & - & - & - \\
AO-Planner~\citep{chen2025affordances} & & \checkmark & \checkmark & & & 5.55 & 59.0 & 47.0 & 33.0 & & 7.06 & 43.3 & 30.5 & 50.1 \\
Seq2Seq~\citep{krantz_vlnce_2020} & \checkmark & & \checkmark & & & 7.77 & 37.0 & 25.0 & 22.0 & & 12.10 & 13.9 & 11.9 & 30.8 \\
CMA~\citep{krantz_vlnce_2020} & \checkmark & & \checkmark & & & 7.37 & 40.0 & 32.0 & 30.0 & & - & - & - & - \\
RGB-Seq2Seq~\citep{krantz_vlnce_2020} & \checkmark & & & & & 10.10 & 8.0 & 0.0 & 0.0 & & - & - & - & - \\
RGB-CMA~\citep{krantz_vlnce_2020} & \checkmark & & & & &  9.55 & 10.0 & 5.0 & 4.0 & & - & - & - & - \\
NaVid~\citep{zhang2024navid} & \checkmark & & & & &  5.47 & 49.0 & 37.0 & 35.0 & & - & - & - & - \\
Uni-NaVid~\citep{zhang2024uninavid} & \checkmark & & & & & 5.58 & 53.5 & 47.0 & 42.7 & & 6.24 & 48.7 & 40.9 & - \\
NaVILA~\citep{cheng2024navila} & \checkmark & & & & & 5.22 & 62.5 & 54.0 & 49.0 & & 6.77 & 49.3 & 44.0 & 58.8 \\
VLN-R1~\citep{qi2025vln} & \checkmark & & & & & 7.00 & 41.2 & 30.2 & 21.8 & & 9.10 & 22.7 & 17.6 & - \\
OctoNav~\citep{gao2025octonav} & \checkmark & & & & & - & 42.9 & 37.1 & 33.6 & & - & - & - & - \\
StreamVLN~\citep{wei2025streamvln} & \checkmark & & & & & \underline{4.98} & \underline{64.2} & \underline{56.9} & \underline{51.9} & & \underline{6.22} & \underline{52.9} & \underline{46.0} & \underline{61.9} \\
\rowcolor{myblue}
\textbf{AwareVLN (Ours)} & \checkmark  & & & & & \bf4.02 & \bf73.5 & \bf65.4 & \bf55.1 & & \bf3.95 & \bf67.6 & \bf56.1 & \bf65.7 \\
\bottomrule
\end{tabular}}}
\vspace{-6pt}
\label{tab:r2r_rxr}
\end{table*}

Through carefully designed prompts that concatenate these multimodal cues, Qwen-VL-Max is able to generate structured reasoning text reflecting both perception and cognition. Each reasoning output consists of three components mentioned above (Scene description, Progress assessment, and Plan for the next step). This automated generation process provides a scalable and reliable source for training the unified reason-act framework with self-aware capability.

\subsection{Training and Inference}
The training of AwareVLN leverages diverse data sources. In the pre-training stage, we follow NaVILA~\cite{cheng2024navila} and incorporate not only common navigation data but also large-scale vision–question answering datasets.
During the fine-tuning stage, we utilize the reasoning-augmented navigation trajectories produced by our Automatic Data Engine, along with additional human videos~\cite{cheng2024navila} without reasoning supervision to improve generalization.
This strategy allows the model to retain strong visual grounding and linguistic alignment while acquiring self-awareness reasoning abilities through targeted supervision. The training of AwareVLN is conducted on four nodes of NVIDIA H20 GPUs. 
During inference, we use an NVIDIA RTX 4090 GPU, achieving an inference speed of roughly 1 FPS.

\section{Experiments}
    In this section, we first describe our experimental setting. Then we compare AwareVLN with state-of-the-art vision-and-language-navigation methods. We also conduct ablation experiments on our framework to validate the effectievness of each design choice. Finally, we provide qualitative analysis through rollout visualizations in both simulated and real-world environments.

\subsection{Experimental Setup}
For quantitative analysis, we compare different mainstream VLN methods on simulation benchmarks.

\textbf{Benchmarks.} We conduct simulator experiments on the VLN-CE benchmarks R2R-CE~\cite{anderson2018vision} and RxR-CE~\cite{ku2020room}. These datasets are obtained by converting the discrete trajectories from the original R2R and RxR datasets into continuous paths within the Habitat simulator~\cite{savva2019habitat}. Their environments are sourced from the MP3D dataset~\cite{Matterport3D}. For evaluation, we use the validation-unseen split, which contains 1,839 episodes for R2R-CE and 11,006 episodes for RxR-CE. R2R-CE instructions are all in English, while RxR-CE includes three languages. Compared with R2R-CE, RxR-CE has longer trajectories and more detailed instructions.

\textbf{Evaluation Metrics.} Following the evaluation setup in~\cite{anderson2018evaluation,ilharco2019general}, we adopt \textit{success rate (SR)} and \textit{success rate weighted by path length (SPL)} as our primary performance metrics. 
SR denotes the percentage of episodes in which the agent arrives within $m$ meters of the target position, with $m = 3$. 
SPL extends SR by further considering the trajectory length, thereby indicating how closely the executed path matches the reference path. 
In addition, we also report navigation error (NE) and Oracle Success Rate (OS) as supplementary metrics.

\subsection{Main Results}
We compare AwareVLN with state-of-the-art vision-and-language navigation methods on the challenging R2R and RxR benchmarks, as shown in Table \ref{tab:r2r_rxr}. Note that different methods may adopt different setting of sensors, so we also list the observation format in table, where S.RGB, Pano., Depth and Odo. mean to monocular RGB image, panoptic RGB images, RGB-D input instead of pure RGB and camera pose of each input image respectively. It is shown that AwareVLN achieves leading performance with pure monocular RGB stream as input, which is simple and deployment-friendly. Apart from high performance, AwareVLN is also explainable through self-aware reasoning, enabling accurate self-correction and re-planning in complex environments.

\begin{table}[t]
\centering
\caption{Real-world evaluation across three environments.}
\vspace{-5pt}
\label{tab:real}
\fontsize{5.6pt}{8.pt}\selectfont
\setlength{\aboverulesep}{1pt}
\setlength{\belowrulesep}{1.2pt}

\begin{tabular}{p{0.7cm}
*{12}{>{\centering\arraybackslash}p{0.18cm}}}
\toprule
\multirow{3}{*}{ }
& \multicolumn{4}{c}{\textbf{Corridor}}
& \multicolumn{4}{c}{\textbf{Home}}
& \multicolumn{4}{c}{\textbf{Office}} \\
\cmidrule(lr){2-5}\cmidrule(lr){6-9}\cmidrule(lr){10-13}
& \multicolumn{2}{c}{\textbf{Simple}} & \multicolumn{2}{c}{\textbf{Complex}}
& \multicolumn{2}{c}{\textbf{Simple}} & \multicolumn{2}{c}{\textbf{Complex}}
& \multicolumn{2}{c}{\textbf{Simple}} & \multicolumn{2}{c}{\textbf{Complex}} \\
\cmidrule(lr){2-3}\cmidrule(lr){4-5}
\cmidrule(lr){6-7}\cmidrule(lr){8-9}
\cmidrule(lr){10-11}\cmidrule(lr){12-13}
& NE$\downarrow$ & SR$\uparrow$
& NE$\downarrow$ & SR$\uparrow$
& NE$\downarrow$ & SR$\uparrow$
& NE$\downarrow$ & SR$\uparrow$
& NE$\downarrow$ & SR$\uparrow$
& NE$\downarrow$ & SR$\uparrow$ \\
\midrule

NaVid
& 2.66 & \underline{0.33} & 4.12 & 0.00
& \underline{1.99} & \underline{0.67} & 2.44 & 0.33
& 2.30 & 0.33 & 3.89 & 0.00 \\
NaVILA
& \underline{2.34} & \underline{0.33} & \underline{2.64} & \underline{0.33}
& 2.17 & \underline{0.67} & \underline{2.32} & \underline{0.67}
& \underline{2.19} & \underline{0.67} & \underline{2.55} & \underline{0.33} \\
\rowcolor{myblue}
AwareVLN
& \textbf{1.86} & \textbf{1.00} & \textbf{2.31} & \textbf{0.67}
& \textbf{1.54} & \textbf{1.00} & \textbf{1.93} & \textbf{1.00}
& \textbf{1.77} & \textbf{1.00} & \textbf{2.26} & \textbf{0.67} \\

\bottomrule
\end{tabular}
\vspace{-6pt}
\end{table}

\subsection{Real-world Evaluation}
To validate the sim-to-real generalization ability, we conduct quantitative evaluations in real-world environments. Specifically, we collect a set of 18 navigation instructions spanning both simple and complex tasks, with varying instruction lengths, across three representative environments: Corridor, Home, and Office.
As shown in Tab.~\ref{tab:real}, our method consistently outperforms all baselines across different environments and task complexities. 
These results demonstrate that the proposed self-aware reasoning mechanism effectively enhances sim-to-real generalization in diverse real-world scenarios.

\subsection{Ablation Study}
We conduct thorough ablation studies to validate the effects of each component in AwareVLN. All experiments are performed on the R2R-CE and RxR-CE Val-Unseen splits.

\textbf{Analysis of Automatic Data Engine.} As shown in Table~\ref{tab:ablation_correctnav}, we analyze the impact of different key reasoning nodes defined in automatic data engine. The complete method, utilizing all critical nodes, achieves the best performance across all metrics. Removing any component leads to a consistent performance drop. Specifically, ablating the \emph{Subtask Completion} causes the most significant performance decline, as the model loses track of its progress within the overall instruction. Removing \emph{Path Deviation} hinders the model's ability to recognize and correct navigation errors, while disabling \emph{Stopping Error} prevents the agent from accurately determining when the goal is reached. These results demonstrate that each element in our reasoning structure is critical for fostering robust self-awareness.

\textbf{Effect of Model Architecture and Reasoning Schedule.} We further investigate the design of our reasoning mechanism in Table~\ref{tab:special_tokens_compare}. Our full model, which uses special tokens to orchestrate a \emph{sparse} reasoning schedule (i.e., reasoning only at keyframes), outperforms two alternative architectures. First, removing the special tokens and forcing the model to directly predict actions or reasoning degrades performance, validating that the structured output facilitated by the tokens is essential for clear task decomposition. Second, the variant that performs reasoning and action prediction densely at \emph{every} frame (\emph{Reason with action densely}) shows a considerable performance drop. This not only confirms that our sparse reasoning is more effective for decision-making but also makes our method highly efficient, as the computationally intensive reasoning process is only triggered when necessary.

\begin{table}[t]
    \small
    \centering
    \caption{Ablation study of different key reasoning nodes defined in automatic data engine on R2R-CE and RxR-CE Val-Unseen splits.}
    \vspace{-5pt}
    \setlength{\tabcolsep}{2.5pt}
    \scalebox{0.99}{
{\fontsize{7pt}{8pt}\selectfont
\begin{tabular}{lccclccc}
\toprule
& \multicolumn{3}{c}{R2R-CE Val-Unseen} & & \multicolumn{3}{c}{RxR-CE Val-Unseen} \\
\cmidrule(lr){2-4} \cmidrule(lr){6-8}
& NE $\downarrow$ & SR $\uparrow$ & SPL $\uparrow$ & & NE $\downarrow$ & SR $\uparrow$ & SPL $\uparrow$ \\
\midrule
\rowcolor{myblue}
\textbf{Complete Reasoning Data} & \bf4.02 & \bf65.4 & \bf55.1 & & \bf3.95 & \bf67.6 & \bf56.1 \\
\hspace{1em} w/o Subtask Completion & 4.92 & 52.3 & 50.7 & & 4.85 & 52.7 & 45.0 \\
\hspace{1em} w/o Path Deviation & 4.70 & 55.1 & 51.5 & & 4.67 & 54.0 & 49.3 \\
\hspace{1em} w/o Stopping Error & 4.76 & 60.0 & 57.5 & & 4.87 & 61.2 & 53.2 \\
\bottomrule
\end{tabular}}}
\vspace{-4pt}
\label{tab:ablation_correctnav}
\end{table}

\begin{table}[t]
    \small
    \centering
    \caption{Comparison of performance with and without special tokens on R2R-CE and RxR-CE Val-Unseen splits.}
    \vspace{-4pt}
    \setlength{\tabcolsep}{3pt}
    \scalebox{0.99}{
{\fontsize{7pt}{8pt}\selectfont
\begin{tabular}{lccclccc}
\toprule
& \multicolumn{3}{c}{R2R-CE Val-Unseen} & & \multicolumn{3}{c}{RxR-CE Val-Unseen} \\
\cmidrule(lr){2-4} \cmidrule(lr){6-8}
& NE $\downarrow$ & SR $\uparrow$ & SPL $\uparrow$ & & NE $\downarrow$ & SR $\uparrow$ & SPL $\uparrow$ \\
\midrule
\rowcolor{myblue}
\textbf{w/ special tokens} & \bf4.02 & \bf65.4 & \bf55.1 & & \bf3.95 & \bf67.6 & \bf56.1 \\
w/o special tokens & 4.60 & 62.5 & 53.3 & & 4.32 & 62.1 & 53.8 \\
Reason with action densely & 4.27 & 63.8 & 54.2 & & 4.25 & 65.4 & 54.9 \\
\bottomrule
\end{tabular}}}
\vspace{-6pt}
\label{tab:special_tokens_compare}
\end{table}

\subsection{Rollout Visualization}
To qualitatively evaluate the self-aware reasoning capability of AwareVLN, we visualize representative navigation rollouts in both simulated (Figure~\ref{fig:vis1}) and real-world (Figure~\ref{fig:vis2}) environments. These cases demonstrate how our model's structured reasoning translates into robust navigation behavior like self-correction when detecting path deviation and re-planning when completing current sub-task.

\begin{figure*}
	\centering
	\includegraphics[width=\linewidth]{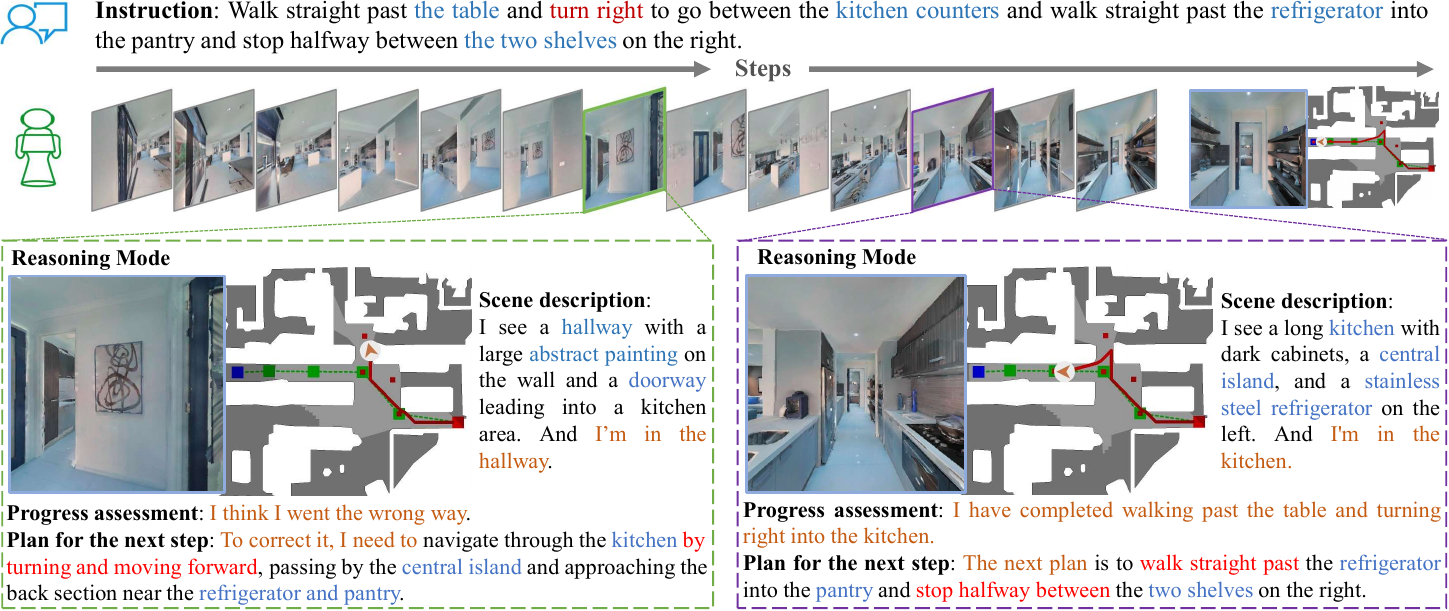}
    \vspace{-.5cm}
    \caption{Rollout in Habitat simulator. AwareVLN performs self-aware reasoning during navigation. As shown in the left part, the agent mistakenly interprets the instruction’s “turn right”. By comparing observations with the instruction, the agent identifies the deviation, and generates a corrective plan. As shown in the right part, after successfully entering the kitchen, AwareVLN recognizes that a subtask has been completed and accurately assesses the navigation progress, producing an appropriate next-step plan aligned with the instruction.}
	\label{fig:vis1}
    \vspace{.2cm}
\end{figure*}

\begin{figure*}
	\centering
	\includegraphics[width=\linewidth]{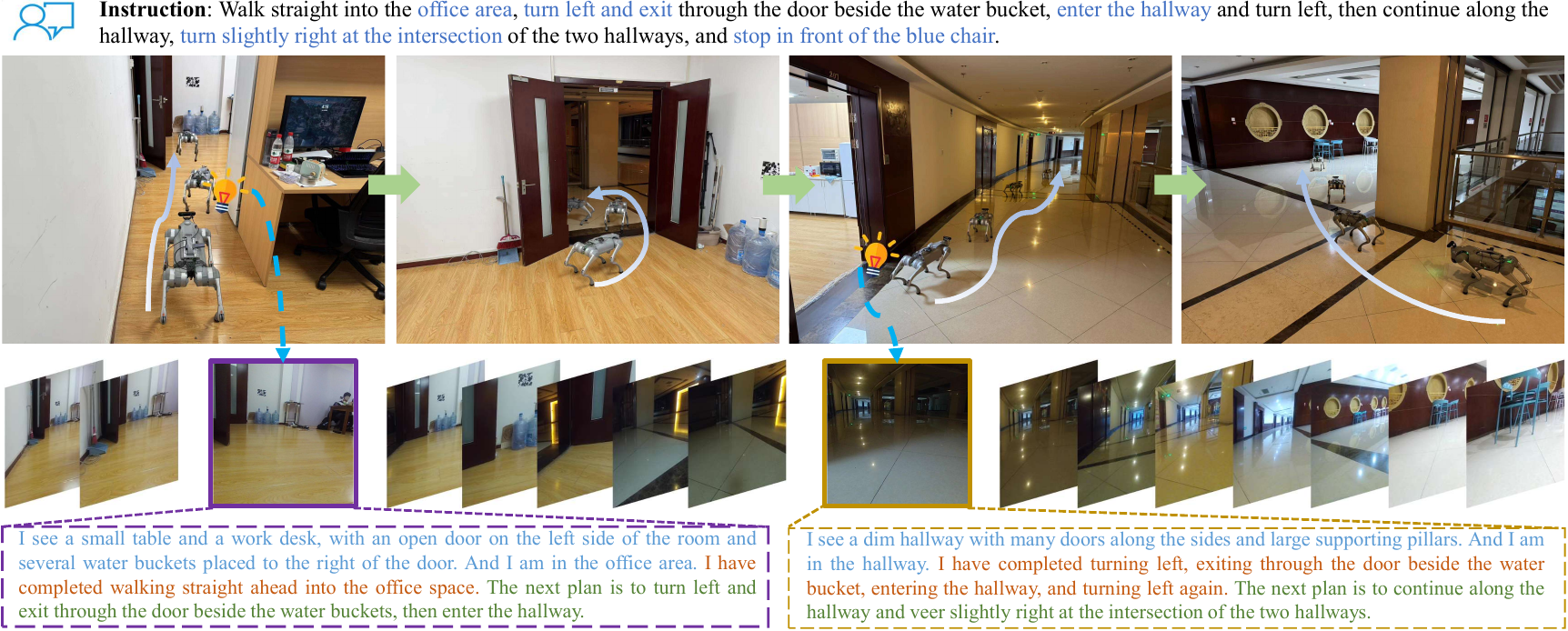}
    \vspace{-.5cm}
    \caption{Rollout in real world. AwareVLN, trained solely on simulator-based navigation data, is deployed on a quadruped robot and successfully completes a long-horizon VLN task in the real world. The agent performs accurate self-aware reasoning at multiple subtask boundaries, demonstrating strong sim-to-real generalization.}
	\label{fig:vis2}
    \vspace{-.4cm}
\end{figure*}

\section{Conclusion}
    In this work, we address a critical limitation in current Vision-and-Language Navigation (VLN) systems: the lack of explicit, explainable reasoning about the agent's state in relation to the instruction and environment. While existing VLM-based methods leverage powerful pre-trained models for direct action prediction, they often operate as black boxes, lacking self-awareness and struggling with error recovery. To bridge this gap, we propose \textbf{AwareVLN}, a novel framework that introduces a self-aware reasoning mechanism into end-to-end navigation. Our approach includes a structural reasoning module that enables the agent to explicitly track its state and task progress, and an automatic data generation strategy with progress division for effective training.
Extensive experiments demonstrate that AwareVLN achieves state-of-the-art performance on standard benchmarks including R2R and RxR. More importantly, our model generates interpretable reasoning traces during navigation in both simulated and real-world environments, validating its self-aware capabilities.

\textbf{Potential Limitation.} Although AwareVLN performs deep reasoning based on observations, we observe that its perception of 3D environments can sometimes be imprecise when deployed on real robots—for example, occasionally bumping into doors or stopping slightly off the intended target. In future work, we plan to explore more robust 3D scene representations derived from monocular RGB inputs to further enhance navigation accuracy.

\section*{Acknowledgement}
This work was supported in part by the National Natural Science Foundation of China under Grant 62321005, Grant 62125603, Grant 62376132, Grant 625B2108, and Grant 624B2076, and in part by the Beijing Natural Science Foundation under Grant No. L247009.

{
    \small
    \bibliographystyle{ieeenat_fullname}
    \bibliography{main}
}

\clearpage
\appendix
\setcounter{page}{1}
\maketitlesupplementary

\begin{figure}[t]
    \centering
    \includegraphics[width=\linewidth]{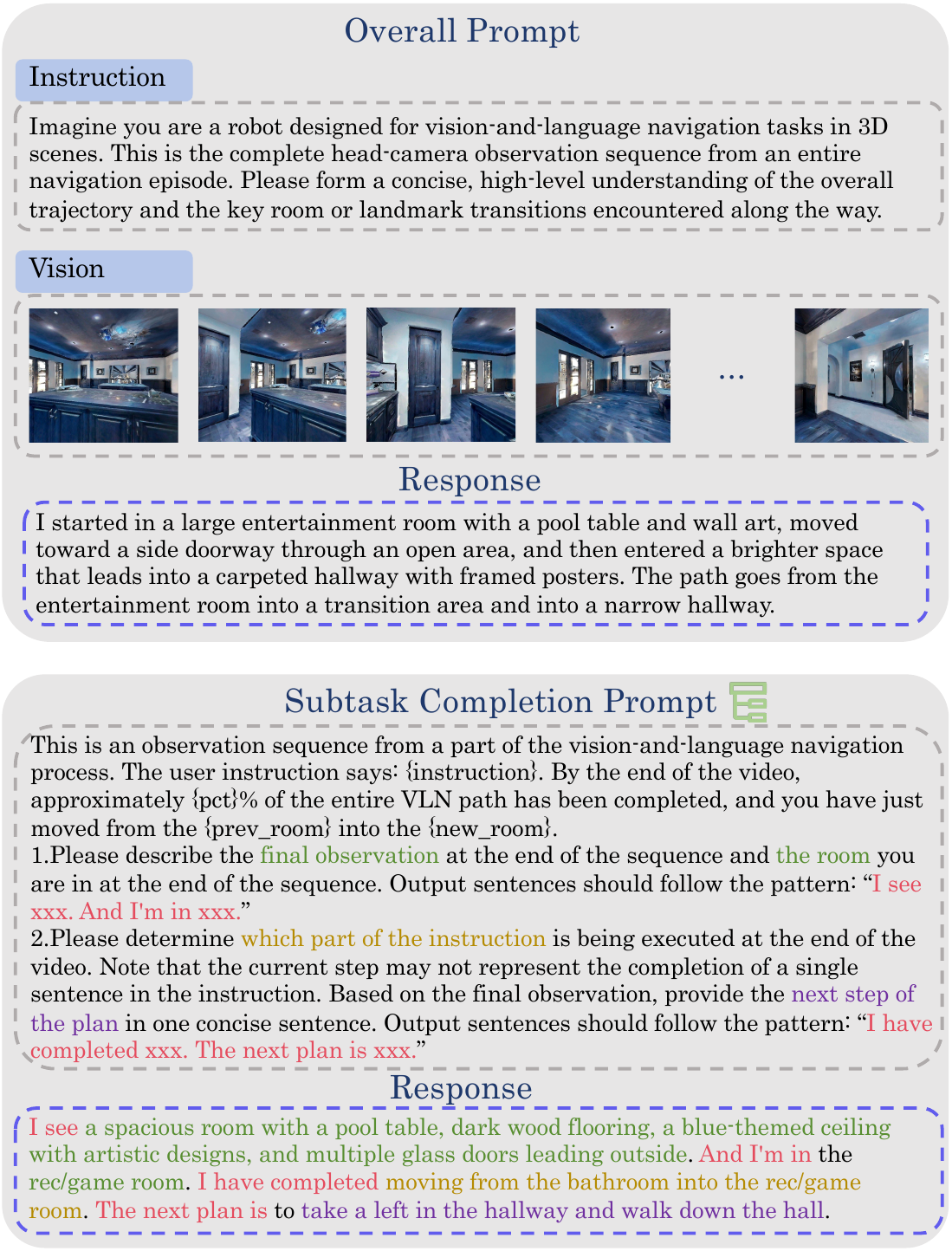}
    \caption{Example of our multi-turn reasoning supervision process (Part 1): global understanding of the navigation episode and reasoning for subtask completion based on localized observations.}
    \label{fig:supp1}
\end{figure}

\begin{figure}[t]
    \centering
    \includegraphics[width=\linewidth]{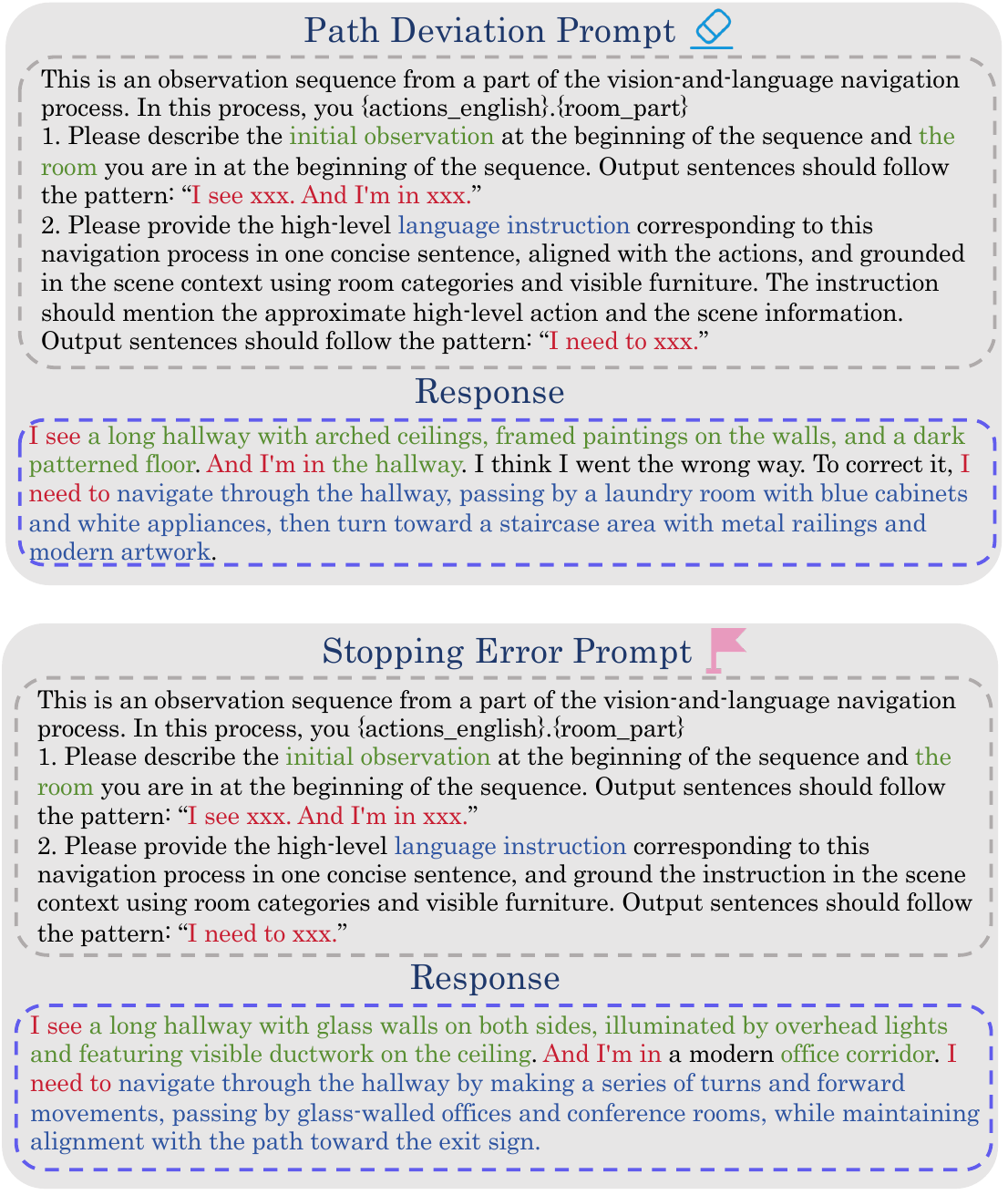}
    \caption{Example of our multi-turn reasoning supervision process (Part 2): reasoning for subsequent node types, including path deviation and stopping error, demonstrating error interpretation and recovery planning.}
    \label{fig:supp2}
\end{figure}

\section{Example of Multi-turn Reasoning Generation}
To further illustrate the effectiveness of our data engine, we provide a concrete example of the multi-turn VLM interaction used for reasoning supervision generation, as shown in Figure~\ref{fig:supp1} and Figure~\ref{fig:supp2}. Starting from the complete episode observation sequence, the VLM first produces a global summary of the navigation trajectory. Then, for each automatically detected key node, we prompt the model with localized visual inputs and node-specific context. The VLM subsequently generates structured reasoning outputs for different cases, including subtask completion, path deviation, and stopping error. This example demonstrates how our prompting design progressively guides the model to understand the navigation state, identify potential issues, and produce reasoning feedback aligned with real navigation progress.

\section{Cross-dataset Generalization}
To further validate the generalization ability of different methods, we conduct a cross-dataset experiment in Table \ref{tab:rxr_cross}. All models are trained exclusively on RxR-CE training set and then evaluated on RxR-CE Val-Unseen split. Our AwareVLN also achieves leading performance under this transfer setting, demonstrating strong robustness.

\begin{table}[h]
    \small
    \centering
    \caption{Cross-dataset performance on the RxR-CE Val-Unseen split. All results are obtained without training on RxR-CE.}
    \vspace{-4pt}
    \resizebox{0.48\textwidth}{!}{%
\begin{tabular}{lccclcccc}
\toprule
& \multicolumn{3}{c}{Observation} & & \multicolumn{4}{c}{RxR Val-Unseen}\\
\cmidrule(lr){2-4} \cmidrule(lr){6-9}
& S.RGB & Depth & Odo. & & NE $\downarrow$ & OS $\uparrow$ & SR $\uparrow$ & SPL $\uparrow$ \\
\midrule
LAW~\cite{raychaudhuri2021language} & \checkmark & \checkmark & \checkmark & & 10.87 & 21.0 & 8.0 & 8.0 \\
CM2~\cite{georgakis2022cross} & \checkmark & \checkmark & \checkmark & & 8.98 & 25.3 & 14.4 & 9.2 \\
WS-MGMap~\cite{chen2022weakly} & \checkmark & \checkmark & \checkmark & & 9.83 & 29.8 & 15.0 & 12.1 \\
 Seq2Seq~\cite{krantz_vlnce_2020} & \checkmark & \checkmark & &
 & 11.8 & 5.02 & 3.51 & 3.43 \\
 CMA~\cite{krantz_vlnce_2020} & \checkmark & \checkmark & & 
 & 11.7 & 10.7 & 4.41 & 2.47 \\
 \midrule
 RGB-Seq2Seq~\cite{krantz_vlnce_2020} & \checkmark & & & &  11.2 & 12.2 & 0.0 & 0.0 \\
 RGB-CMA~\cite{krantz_vlnce_2020} & \checkmark & & & 
 & 9.55 & 14.8 & 0.0 & 0.0 \\
 A$^{2}$NAV~\cite{chen20232} & \checkmark & & & 
 & - & - & 16.8 & 6.3 \\
 NaVid~\cite{zhang2024navid} & \checkmark & & & 
 & \underline{8.41} & 34.5 & 23.8 & 21.2 \\
 NaVILA & \checkmark & & & 
 & 8.78 & \underline{46.8} & \underline{34.3} & \underline{28.2} \\
\rowcolor{myblue}
\textbf{AwareVLN (Ours)} & \checkmark & & & 
 & \bf7.15 & \bf51.0 & \bf39.8 & \bf36.0 \\
\bottomrule
\end{tabular}}
\vspace{-4pt}
\label{tab:rxr_cross}
\end{table}

\section{Visualization of Automatically Collected Trajectories}
Figure~\ref{fig:vis1} and Figure~\ref{fig:vis2} present two representative navigation trajectories collected by our automatic data engine. Each example follows a natural language instruction and illustrates the generated reasoning supervision at key navigation nodes. The visual observations, together with the VLM outputs, clearly demonstrate the agent's understanding of the correction process, subtask completion, and next-step planning, validating the effectiveness of our data engine in producing scalable and interpretable supervision. 
Furthermore, the examples highlight how the agent analyzes navigation status. This showcases not only its ability to align reasoning with navigation progress but also the robustness of the data engine in capturing realistic behaviors and generating fine-grained supervision signals that benefit training self-aware navigation models.

\begin{figure*}[t]
    \centering
    \includegraphics[width=\linewidth]{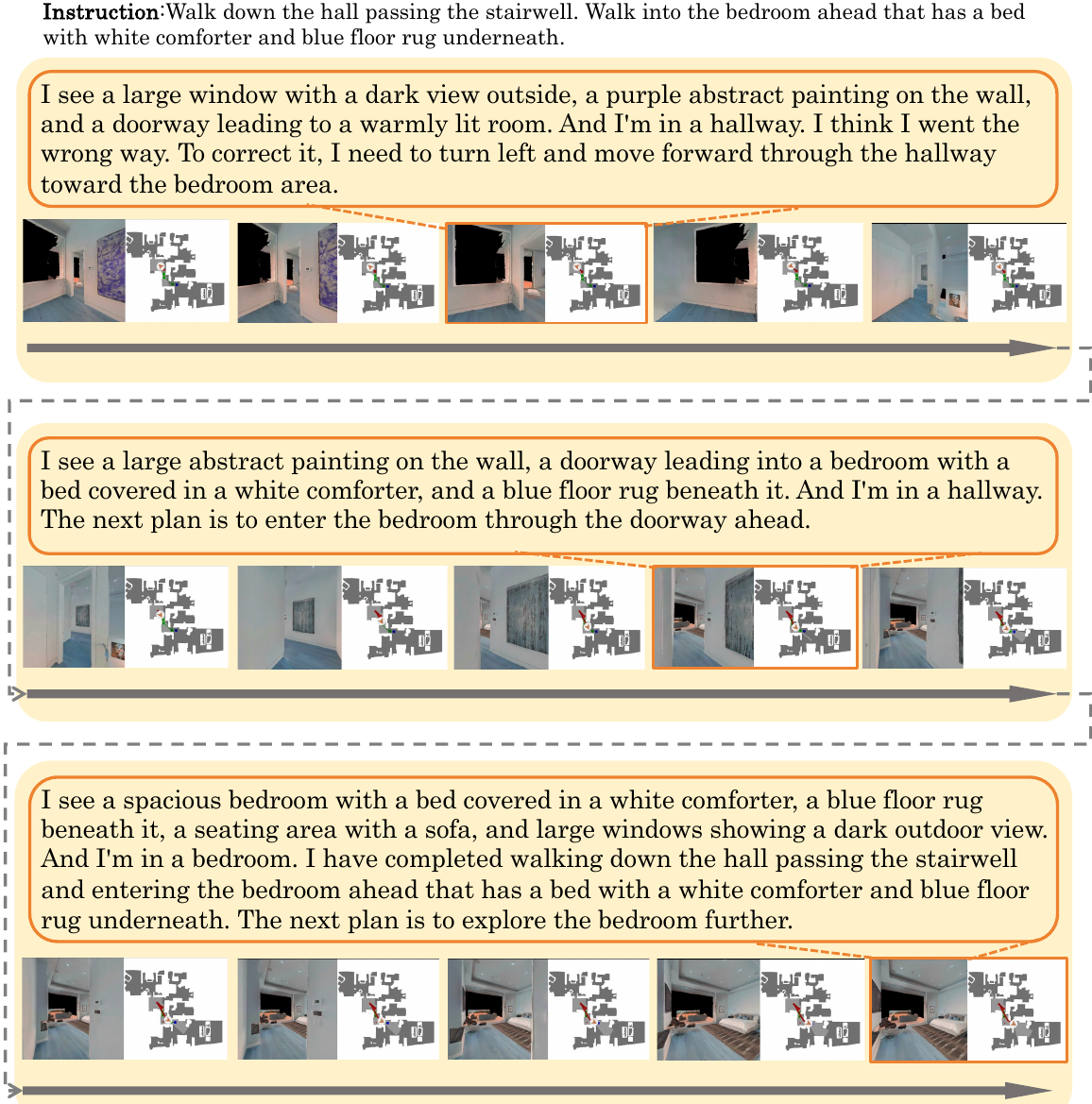}
    \caption{Example 1 of an automatically collected training trajectory, illustrating three key nodes: path deviation, correction completion, and subtask completion. The agent interprets the evolving visual scene and progressively generates structured reasoning outputs aligned with the navigation instruction.}
    \label{fig:vis1}
\end{figure*}

\begin{figure*}[t]
    \centering
    \includegraphics[width=0.95\linewidth]{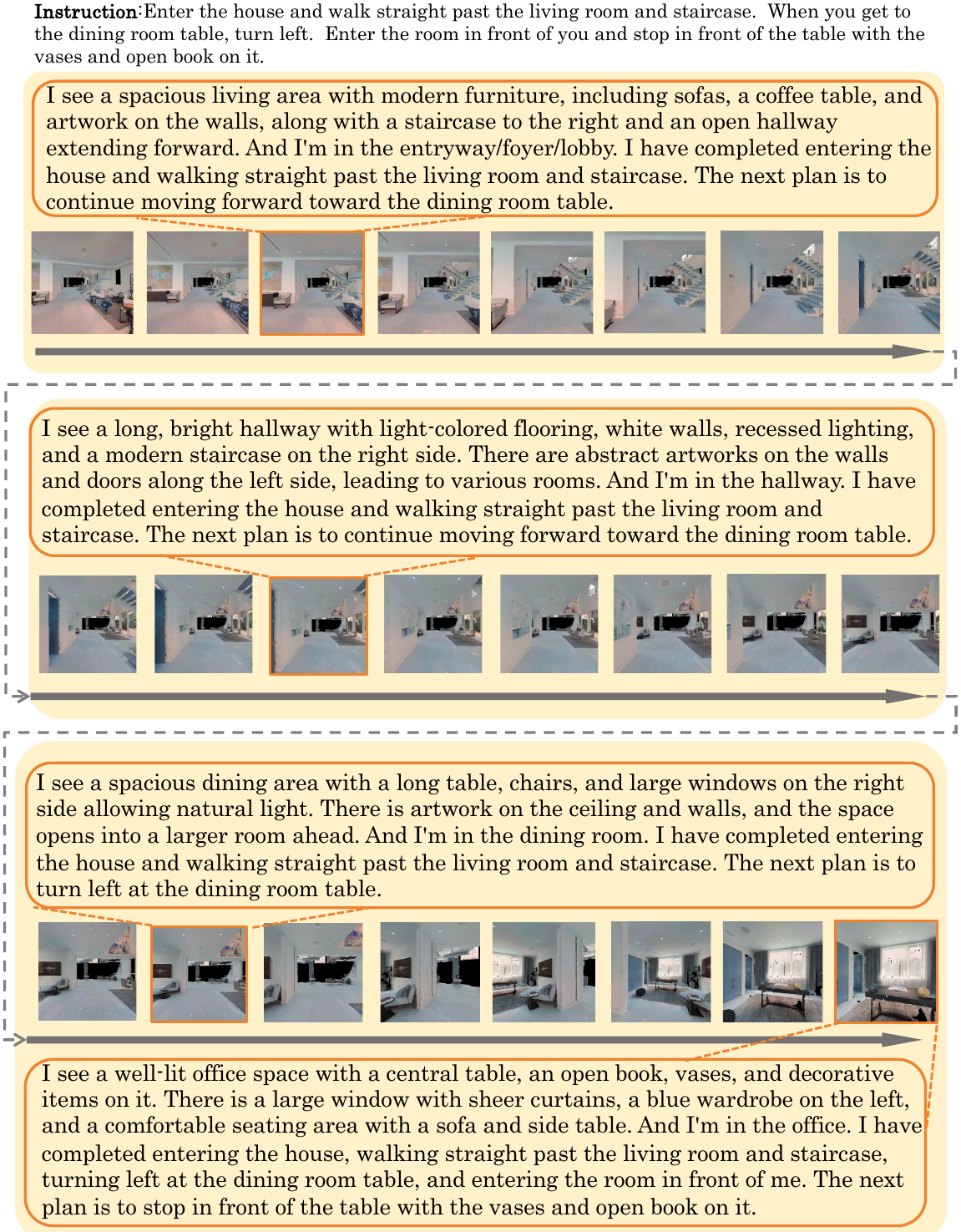}
    \caption{Example 2 of an automatically collected training trajectory. This case includes multiple subtask completion nodes and room transitions, demonstrating the model's ability to reason over spatial changes and identify subtask boundaries during navigation.}
    \label{fig:vis2}
\end{figure*}

\end{document}